\documentclass[runningheads,a4paper]{llncs}
\usepackage{llncsdoc}
\usepackage{gensymb}
\usepackage{graphicx}
\usepackage{verbatim}
\usepackage{wrapfig}
\usepackage[table,xcdraw]{xcolor}
\usepackage{multirow}
\usepackage{amssymb}
\usepackage{rotating}
\usepackage{adjustbox}
\usepackage[utf8]{inputenc}
\usepackage{url}
\usepackage{amsmath}
\usepackage{enumitem}
\usepackage{tikz}
\usetikzlibrary{arrows}
\usepackage[T1]{fontenc}
\usepackage[utf8]{inputenc}
\usepackage{tabularx,ragged2e,booktabs,caption}

\graphicspath{{images/}}
\begin{document}

\title{Simultaneous Detection and Quantification of Retinal Fluid with Deep Learning}
\titlerunning{Deep Learning for Retinal Fluid Quantification}

\author{Dustin Morley, Hassan Foroosh, Saad Shaikh, Ulas Bagci}

\authorrunning{ }

\institute{University of Central Florida, Orlando FL, 32816\\
Email: dustinrmorley@knights.ucf.edu}
\maketitle

\begin{abstract}
We propose a new deep learning approach for automatic detection and segmentation of fluid within retinal OCT images. The proposed framework utilizes both {\it ResNet} and {\it Encoder-Decoder} neural network architectures. When training the network, we apply a novel data augmentation method called {\it myopic warping} together with standard rotation-based augmentation to increase the training set size to $45$ times the original amount. Finally, the network output is post-processed with an energy minimization algorithm (graph cut) along with a few other knowledge guided morphological operations to finalize the segmentation process. Based on OCT imaging data and its ground truth from the RETOUCH challenge, the proposed system achieves dice indices of $0.522$, $0.682$, and $0.612$, and average absolute volume differences of $0.285$, $0.115$, and $0.156$ mm$^3$ for intaretinal fluid, subretinal fluid, and pigment epithelial detachment respectively.
\end{abstract}

\section{Introduction}
Automatic detection and segmentation of fluid within retinal optical coherence tomography (OCT) images is a  task of great importance to the field of ophthalmology. Fluid is not normally present in the retina and its presence decreases visual acuity thus mandating therapeutic intervention. Three types of fluid occur in the retina: intaretinal fluid (IRF), subretinal fluid (SRF), and pigment epithelial detachment (PED). 

To automatically and simultaneously detect and quantify these fluid types, we propose a deep learning based algorithm. Toward this end, we constructed a convolutional neural network (CNN) which takes as input a single $xy$-plane slice from an OCT image and produces a map showing the probabilities of each pixel containing each fluid type as output. We also designed a post-processing framework centered on the graph cut algorithm to produce a final segmentation from the CNN output. \\

\noindent\textbf{Related works.} Deep learning is currently revolutionizing many fields of automated image analysis \cite{GirshickRich,FCN}, and recent advances in GPU hardware alongside novel algorithms have made it possible to apply these methods to medical imaging. In our context, the most significant recent non-hardware development is the use of deconvolution layers to perform bilinear upsampling within a CNN~\cite{FCN}, which allows the output to be the same size as the input despite the use of subsampling operations in the CNN.

Prior published work dealing with simultaneous detection and segmentation of IRF, SRF, and PED in OCT images of the human retina can be found in~\cite{Chen3D,DolejsiSemi}. There are also studies dealing with binary detection of either fluid in general or only one specific type of fluid~\cite{Quellec3D,SunAutomated}. Our method is closest to that of~\cite{Chen3D}, the differences being our use of a deep CNN instead of their initialization method along with simpler post-processing methodology. Interestingly, we demonstrate good performance without utilizing a retinal layer segmentation.

\section{Method}
Our method for simultaneous detection and segmentation of fluid is centered on the use of a deep CNN to assign correct labels to individual OCT slices. Prior to training or using the CNN, images must be standardized by a set of pre-processing steps. Similarly, post-processing steps are utilized after CNN inference in order to stitch together the final output and compute the volume of detected fluids. 

\subsubsection{2.1 Pre-Processing.}
We designed a pre-processing framework to prepare the imaging data prior to applying the CNN, which operates as follows. First, each OCT volume is smoothed with a three-dimensional Gaussian kernel. Next, since our CNN takes individual OCT slices as input, $xy$-plane slices are extracted from each smoothed volume and each reference standard volume. As the slices are extracted, the intensities are rescaled to allow them to be saved as standard 8-bit images. Once the slices are extracted, they are resized to a standard size. Since the {\it Heidelberg} slices were the smallest in the data set, their size (512x496) defined the standard. We used bicubic downsampling to resize the images, and nearest neighbor downsampling (out of necessity) to resize the reference standard slices. After resizing, the slices are cropped in the vertical dimension to an area containing the retina with minimal background. In particular, the cropping is to the 512x256 rectangle with the highest intensity sum. This method was validated to always capture the full retina. Finally, the means and standard deviations are normalized for every image. 

\label{sec:pre-processing}

\subsubsection{2.2 Data Augmentation.}
We performed data augmentation to increase the amount of training data to $45$ times the provided amount. Specifically, we utilized rotations in increments of $2\degree$ from $-8\degree$ to $8\degree$, and an original method that we call "myopic warping." Myopic warping involves introducing centralized downward curvature on the entire retina. In order to induce this effect, we warp the image according to an inverse square force emanating from a point some vertical distance away from the center of the image, i.e. $\overrightarrow{v} = \frac{F \hat{r}} {r^2}$ 
where $\overrightarrow{v}$ is the warp vector for a particular pixel, $F$ is the strength of the force field, and $\overrightarrow{r}$ is the vector pointing from the center of the force field to that pixel (rescaled based on the image size to make the tunable parameters more intuitive to work with). There are thus two parameters that govern the warping: $F$ and the vertical location of the force field center. Changes to either parameter in isolation increases or decreases the amount of warping. Increasing $F$ and the center distance simultaneously results in a warping that is more of a downward translation with very little curvature change, while decreasing both results in a curvier retina. Both myopic warping and rotation result in zero-padding in some areas close to the image boundary in order to preserve the size, with the largest such areas occurring above the retina (due to the myopic warping). To prevent this from introducing strong artificial edges, we replaced these areas with an intensity profile similar to the background profile in the image. To do this, we run a 50x10 rectangle across the top of the image, and identify the placement of this rectangle corresponding to the minimum intensity sum within the rectangle. The mean and standard deviation of this image patch are then computed. These are subsequently used to define a normal distribution from which to draw intensity values for filling the zero-padded regions (higher-than-expected standard deviations are reduced to $2$, to protect against cases where it is not possible to find a 50x10 patch that does not contain any retina pixels). The regions are filled in a "blocky" manner - each randomly drawn intensity is used to set the pixel values over an area as large as 13x13. The image is lightly smoothed after all of these operations to restore continuity. Some examples of myopic warping are shown in Figure 1. 

\begin{figure}
\includegraphics[scale=0.16]{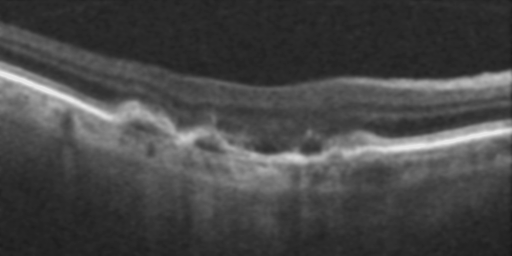}
\includegraphics[scale=0.16]{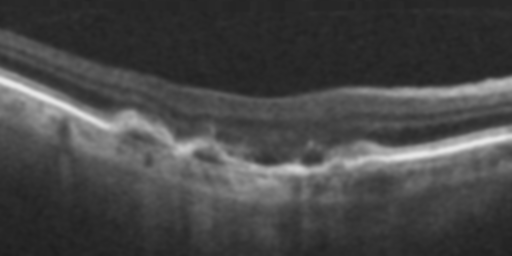}
\includegraphics[scale=0.16]{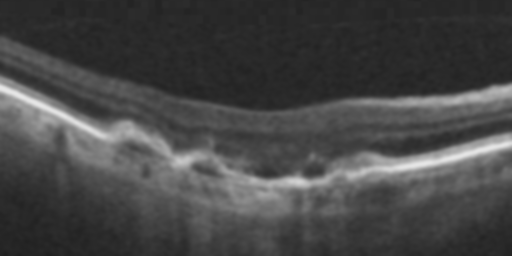}
\includegraphics[scale=0.16]{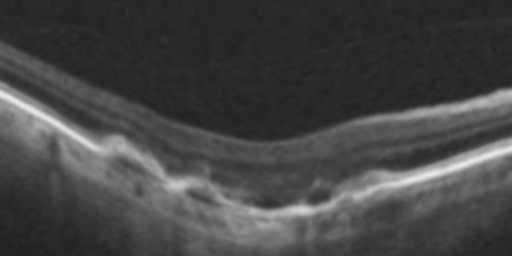}
\caption{Examples of myopic warping. From left to right: original image; applied baseline warping; increased the strength parameter value; decreased the center distance. }
\end{figure}
\vspace{-1cm}
\subsubsection{2.3 CNN Architecture.}
Our CNN for pixelwise segmentation takes a {\it ResNet} approach, utilizing many "skip" layers. The data undergoes a total of three downsampling operations, and is ultimately restored to the original size by three bilinear upsampling layers. A total of 43 convolution layers are contained within the CNN. 32 of these are on the encoder side, and the 3 final convolution layers on the decoder side only contain 4 filters apiece as they are part of a special endgame approach we took. It is of course necessary that the final layer contain only 4 filters for a 4-class labeling problem, but we utilized three such layers to allow the net to learn a basic ``intensity multiplier" for each class with which to amend an initial classification. This is illustrated in Figure \ref{fig:endgame}. As a general rule, all encoder convolution layers were initialized according to the {\it Xavier} scheme, while decoder layers were initialized to zeros instead. 

\begin{figure}[h]
\includegraphics[scale=0.22]{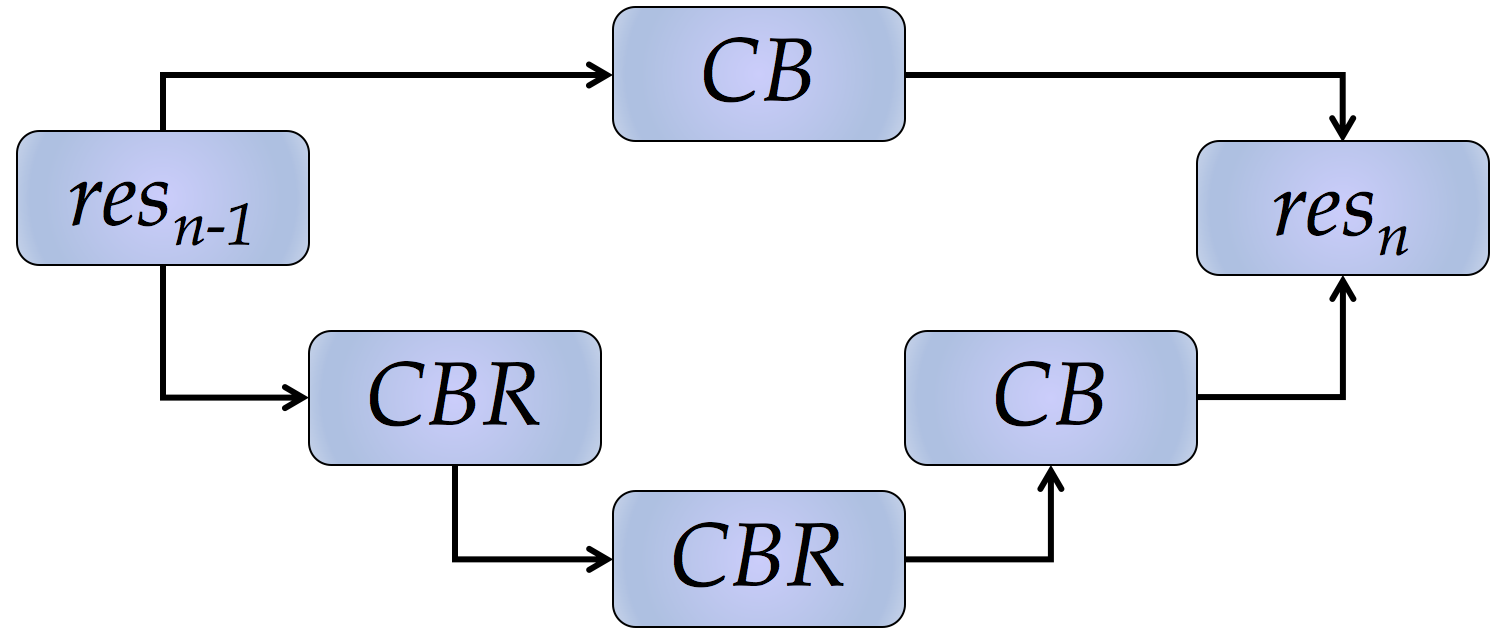}
\includegraphics[scale=0.23]{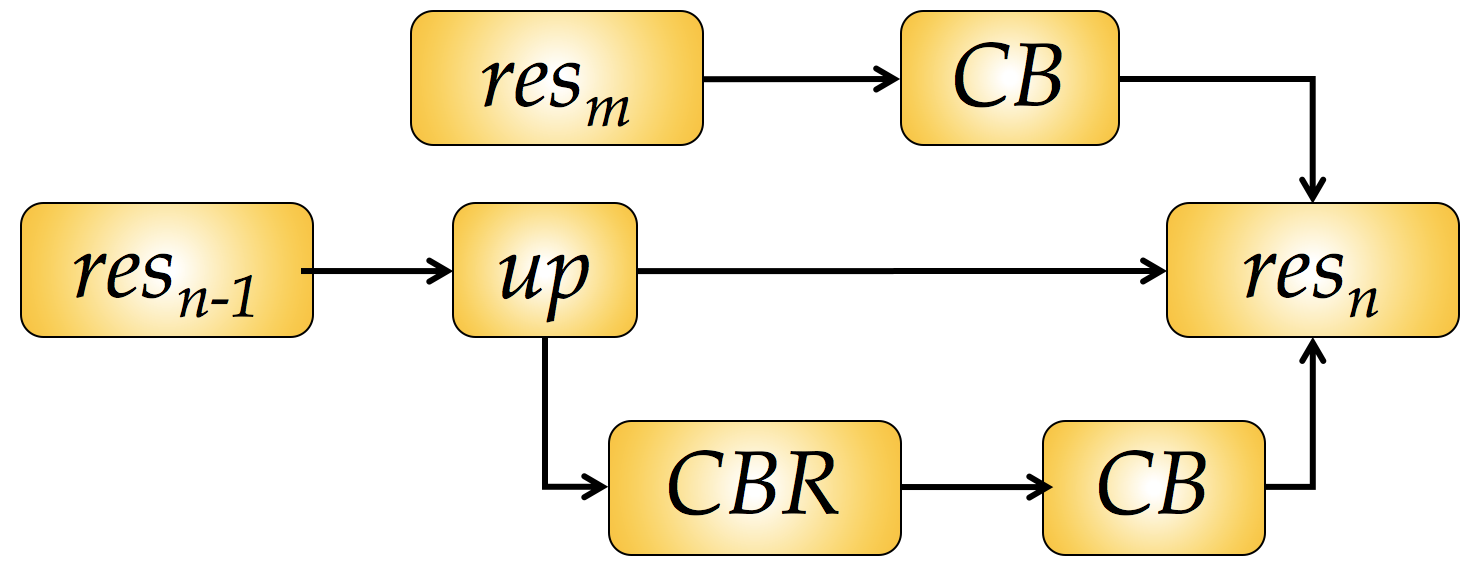}
\caption{Fundamental processing units on the encoder (left, blue) and decoder (right, orange) portions of our CNN. ``CBR" refers to the following full sequence: convolution, BN, ReLU. Similarly, ``CB" is convolution and BN without ReLU. The "res" units perform elementwise addition followed by ReLU. Numbers of filters vary, but the encoder filter sizes are all $11$x$11 \rightarrow 7$x$7 \rightarrow 1$x$1$ for the 3-layer branch and 1x1 for the other branch, while the decoder filter sizes  are all $11$x$11 \rightarrow 7$x$7 $ for the 2-layer branch and $3$x$3$ for the 1-layer branch. Some encoder units also utilize a third branch from an arbitrary earlier "res" unit, with or without passing through a $1$x$1$ convolution layer.\label{fig:encUnit}}
\end{figure}

\begin{figure}[!h]
\tikzstyle{int}=[draw, fill=blue!20, minimum size=2em]
\tikzstyle{init} = [pin edge={to-,thin,black}]

\centering
\begin{tikzpicture}[node distance=1.25cm,auto,>=latex']
	\node [int] (dec) at (-1.5,-0.5) {dec};
    \node [int] (precls) at (0,-0.5) {conv};
    \node [int] (data) at (-1.5,0.5) {data};
    \node [int] (wrap) at (0,0.5) {conv};
    \node [int] (mult) at (1.5,0) {mult};
    \node [int] (res) at (3,0) {res};
    \node [int] (final) at (4.5,0) {conv};
    \node [int] (softmax) at (6,0) {softmax};
    \path[->] (data) edge node {} (wrap);
    \path[->] (dec) edge node {} (precls);
    \path[->] (wrap) edge [bend left = 15] (mult);
    \path[->] (precls) edge [bend right = 15] (mult);
    \path[->] (mult) edge node {} (res);
    \path[->] (precls) edge [bend right = 25] (res);
    \path[->] (res) edge node {} (final);
    \path[->] (final) edge node {} (softmax);
\end{tikzpicture}
\caption{Endgame for the CNN. All convolution layers shown only have 4 filters, with the layer connected to data having $3$x$3$ filters and the others $1$x$1$. The intuition behind this design was to allow the net to easily learn basic intensity rules to be applied on top of the deep features. The convolution layer responsible for learning this logic is given a reduced learning rate, and it contains a ReLU activation. }
\label{fig:endgame}
\end{figure}
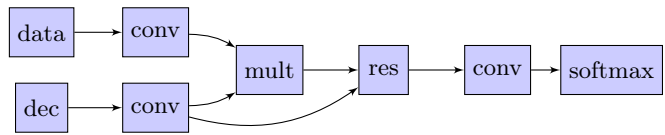

Figure \ref{fig:encUnit} shows the fundamental encoder and decoder $ResNet$ computational units that were utilized. Note that all convolution layers outside the endgame region are followed by batch normalization (BN), and several are additionally followed by a Rectified Linear Unit (ReLU) activation. Table~\ref{table:Arch} contains the full specification of the CNN, broken down unit-by-unit. We trained the CNN for $4$ epochs on our augmented data set, using stochastic gradient descent with a batch size of $8$, a momentum of $0.9$, a weight decay of $5$x$10^{-4}$, and an initial learning rate of $10^{-4}$ which is divided by $10$ after the first two epochs. The network was trained in {\it Caffe} \cite{caffe} using the {\it Infogain} loss function to assign lower weight to non-fluid pixels to balance out the large number of these pixels in relation to fluid pixels. We used two-fold cross-validation, and training on each of the two subsets took roughly $8$ hours on an NVIDIA Titan Xp GPU. Only the central third of $xy$-plane slices from each image volume was used for training, resulting in roughly $1,100$ slices per training subset (roughly $50,000$ after data augmentation).

\vspace{-0.5cm}

\begin{table}[ht]
\centering
\caption{Complete architecture specification for our deep {\it ResNet} encoder-decoder CNN. As an example of how to read this table, Res1 receives data from two parallel branches originating from Res0: one passes through 3 layers of $24,48,$ and $48$ filters, and the other passes through 1 layer of $48$ filters. The filter sizes are all as specified in the descriptions of Figures~\ref{fig:encUnit} and \ref{fig:endgame}.}
\begin{tabular}{p{2.9 cm}p{2.9 cm}p{2 cm}p{1.8cm}p{2cm}}
\hline\noalign{\smallskip}
{\scriptsize \textbf{Computational Unit}} & {\scriptsize \textbf{Source 1}} & {\scriptsize \textbf{Source 2}} & {\scriptsize \textbf{Source 3}} & {\scriptsize \textbf{Output Size}} \\
\noalign{\smallskip}
\hline
\noalign{\smallskip}
{\scriptsize Data}   & {\scriptsize N/A} & {\scriptsize N/A} & {\scriptsize N/A} & {\scriptsize 1x512x256}\\
 {\scriptsize Conv1 (11x11)} & {\scriptsize Data} & {\scriptsize N/A} & {\scriptsize N/A} & {\scriptsize 32x512x256}\\
 {\scriptsize Conv2 (7x7)} & {\scriptsize Conv1} & {\scriptsize N/A} & {\scriptsize N/A} & {\scriptsize 32x256x128} \\
 {\scriptsize Res0} & {\scriptsize Conv2 ($24,48,48$)}  & {\scriptsize Conv2 ($48$)} & {\scriptsize N/A} & {\scriptsize 48x256x128} \\
 {\scriptsize Res1} & {\scriptsize Res0 ($24,48,48$)} & {\scriptsize Res0 ($48$)} & {\scriptsize N/A} & {\scriptsize 48x128x64} \\
 {\scriptsize Res2} & {\scriptsize Res1 ($32,32,48$)} & {\scriptsize Res1 ($48$)} & {\scriptsize Res1} & {\scriptsize 48x128x64} \\
 {\scriptsize Res3} & {\scriptsize Res2 ($32,32,64$)} & {\scriptsize Res2 ($64$)} & {\scriptsize N/A} & {\scriptsize 64x128x64} \\
 {\scriptsize MaxPool} & {\scriptsize Res3} & {\scriptsize N/A} & {\scriptsize N/A} & {\scriptsize 64x64x32} \\
 {\scriptsize Res4} & {\scriptsize MaxPool ($48,48,64$)} & {\scriptsize MaxPool ($64$)} & {\scriptsize N/A} & {\scriptsize 64x64x32} \\
 {\scriptsize Res5} & {\scriptsize Res4 ($48,48,100$)} & {\scriptsize Res4 ($100$)} & {\scriptsize N/A} & {\scriptsize 100x64x32} \\
 {\scriptsize Res6} & {\scriptsize Res5 ($128,64,128$)} & {\scriptsize Res5 ($128$)} & {\scriptsize Res4 ($128$)} & {\scriptsize 128x64x32}\\
 {\scriptsize ConvMid (1x1)} & {\scriptsize Res6} & {\scriptsize N/A} & {\scriptsize N/A} & {\scriptsize 36x64x32} \\
 {\scriptsize Up1} & {\scriptsize ConvMid} & {\scriptsize N/A} & {\scriptsize N/A} & {\scriptsize 36x128x64} \\
 {\scriptsize Res7} & {\scriptsize Up1 ($32,36$)} & {\scriptsize Up1} & {\scriptsize Res3 ($36$)} & {\scriptsize 36x128x64} \\
 {\scriptsize Up2} & {\scriptsize Res7} & {\scriptsize N/A} & {\scriptsize N/A} & {\scriptsize 36x256x128} \\
 {\scriptsize Res8} & {\scriptsize Up2 ($24,36$)} & {\scriptsize Up2} & {\scriptsize Res0 ($36$)} & {\scriptsize 36x256x128} \\
 {\scriptsize Up3} & {\scriptsize Res8} & {\scriptsize N/A} & {\scriptsize N/A} & {\scriptsize 36x512x256} \\
 {\scriptsize Res9} & {\scriptsize Up3 ($16,36$)} & {\scriptsize Up3} & {\scriptsize N/A} & {\scriptsize 36x512x256} \\
 {\scriptsize Endgame (Fig.\ref{fig:endgame})} & {\scriptsize Res9 ($4$)} & {\scriptsize Data ($4$)} & {\scriptsize N/A} & {\scriptsize 4x512x256} \\
\hline
\label{table:Arch}
\end{tabular}
\end{table}

\vspace{-0.5cm}
\subsubsection{2.4 Post-Processing.}
We utilized multiple post-processing algorithms to improve upon the CNN output before constructing the final output. Central to the post-processing is the graph-cut algorithm~\cite{GC1,GC2}. We utilized a MATLAB wrapper~\cite{GC4} of the Boykov-Kolmogorov graph cut implementation. Prior to graph cut, we zeroed out IRF probabilities on edge pixels (based on Difference-of-Gaussians (DoG)) and modestly decreased SRF and PED probabilities on bright pixels in continuous fashion, according to equation \ref{eqn:SRF_PED_Reduce}. Specifically, we used $T_1(\mu(I),\sigma(I))=\mu(I)+\sigma(I)$ and $T_2(\mu(I),\sigma(I))=\mu(I)+3\sigma(I)$, with $\lambda=$ $0$ for SRF and $0.95$ for PED.

\begin{equation}
P(x,y)=\left\{\begin{array}{lr}P^\prime (x,y), & I(x,y) \leq T_1(\mu (I), \sigma(I)) \\
P^\prime (x,y)\max(\lambda,\frac{T_2(\mu(I),\sigma(I))-I(x,y)}{T_2(\mu(I),\sigma(I)) - T_1(\mu(I),\sigma(I))}), & \text{otherwise.}
\end{array} \right.
\label{eqn:SRF_PED_Reduce}
\end{equation}

These operations define the prior class probabilities used by the graph-cut algorithm. The data cost was set to the negative logarithm of the prior. The base smoothness cost (penalty for neighboring pixels having different labels) was set to $5$ for IRF/non-fluid, $10$ for all other different label combinations, and $0$ for adjacent pixels having the same label. This cost was then multiplied by a spatially varying smoothness cost, set from the result of applying a DoG filter to the image. In particular, the full smoothness cost is specified in equation \ref{eqn:SmoothCost}, with $S^\prime$ the base smoothness cost and $g(I)$ equal to the result of the DoG operation.

\begin{equation}
S=S^\prime\exp(-5g(I)/\max(g(I))).
\label{eqn:SmoothCost}
\end{equation}

 After graph cut, we invoked two additional post-processing steps. The first enforces the rule that PED cannot occur above IRF or SRF. The approach here is very straightforward. For each vertical line within which PED and either IRF or SRF were contained, the topmost PED pixel was found, and then the topmost IRF or SRF pixel beneath the topmost PED pixel was found. Counts were obtained for the number of pixels belonging to PED and the number of pixels belonging to the other identified fluid beneath the first PED occurrence. The larger count ``wins," meaning that all pixels of the "losing" fluid have their labels replaced by the "winning" fluid. The final post-processing step here was some PED connected component analysis, which simply removed PED connected components that didn't meet criteria for a minimum slope change across the top (the logic being that the top of a PED occurrence is never a straight line). 
 
These steps resulted in each slice of the OCT volume having been fully processed in its own right, but without leveraging any 3D information; obviously, there should be reasonable agreement between adjacent slices of the same OCT volume. To leverage this, we built the result volumes and then ran graph cut on all of the $yz$-plane slices. The result volumes constructed at this stage were "uncropped" back to the standard size of 512x496, but were not resized to the original image sizes until after running graph cut on the $yz$-plane slices (see section \ref{sec:pre-processing}). For this graph cut, the smoothness cost was set the same way as described earlier for the $xy$-plane results, except a different parameterization for the DoG filter was used. The data cost at each pixel was zero for the current label at that pixel and a positive constant for the other three classes.

\section{Results}
We evaluated our method by computing dice index and absolute volume difference (AVD), alongside a qualitative evaluation through visual inspection. The results were generated using two-fold cross-validation, with the two subsets having roughly equal amounts of each fluid type and roughly an equal number of scans from each device. 

\begin{figure}[h]
\centering
\includegraphics[width=2.9cm,height=1.2cm]{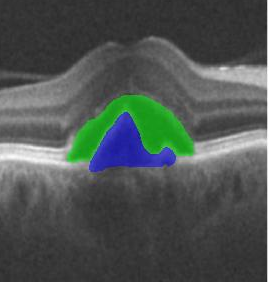}
\includegraphics[width=2.9cm,height=1.2cm]{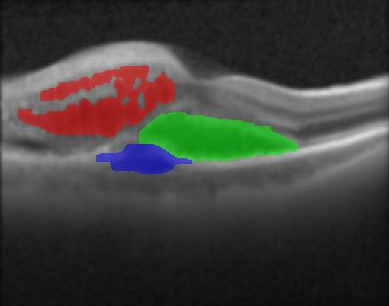}
\includegraphics[width=2.9cm,height=1.2cm]{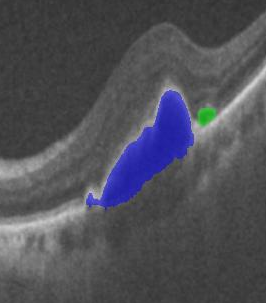}
\includegraphics[width=2.9cm,height=1.2cm]{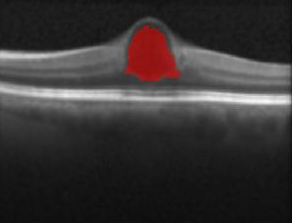}
\linebreak
\includegraphics[width=2.9cm,height=1.2cm]{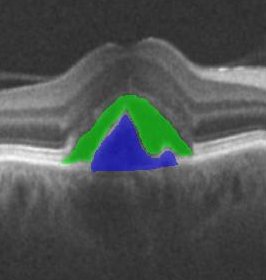}
\includegraphics[width=2.9cm,height=1.2cm]{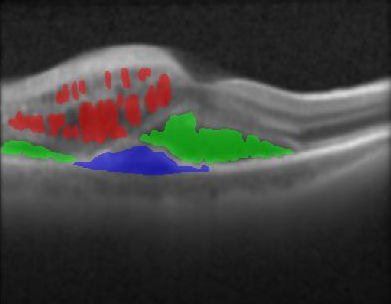}
\includegraphics[width=2.9cm,height=1.2cm]{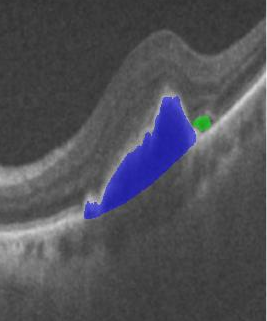}
\includegraphics[width=2.9cm,height=1.2cm]{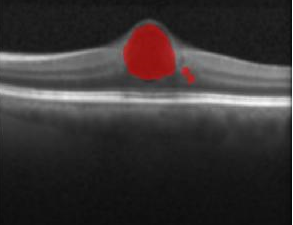}
\caption{Examples on which our method performed extremely well. The top row: the output of our method. Bottom row: the reference standard (Red = IRF, green = SRF, blue = PED). \label{fig:goodEx} }

\begin{tabular}{ c c | c c }
\includegraphics[width=2.9cm,height=1.2cm]{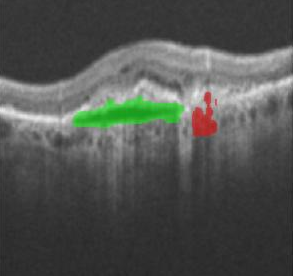} &
\includegraphics[width=2.9cm,height=1.2cm]{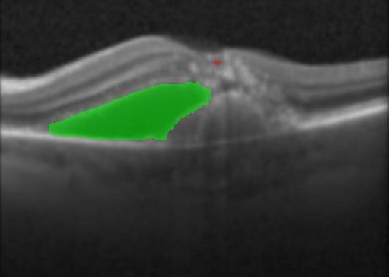} &
\includegraphics[width=2.9cm,height=1.2cm]{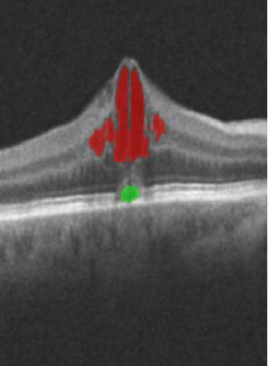} &
\includegraphics[width=2.9cm,height=1.2cm]{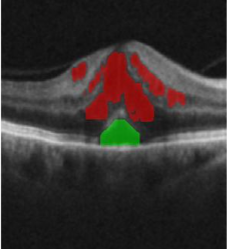}
\\
\includegraphics[width=2.9cm,height=1.2cm]{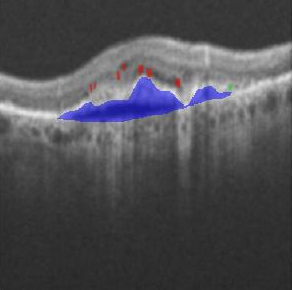} &
\includegraphics[width=2.9cm,height=1.2cm]{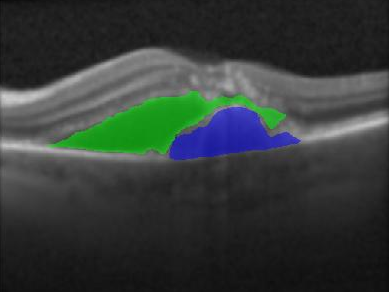} &
\includegraphics[width=2.9cm,height=1.2cm]{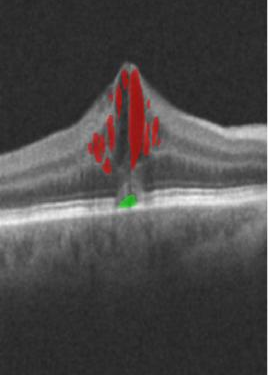} &
\includegraphics[width=2.9cm,height=1.2cm]{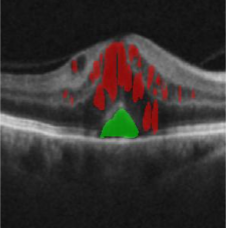}
\\
\multicolumn{2}{c}{\bf{a}} & \multicolumn{2}{c}{\bf{b}}
\end{tabular}
\caption{(a) Examples on which our method struggled, and (b) examples on which our IRF results are arguably more accurate than the reference standard. Top row is the output of our method, and the bottom row is the reference standard (Red = IRF, green = SRF, blue = PED).\label{fig:chalEx} }
\end{figure}

Qualitatively, our method was observed to be capable of obtaining very good results, as visually verified by the participating ophthalmologist, but there were also some challenging cases. Some examples are shown in Figures \ref{fig:goodEx} and \ref{fig:chalEx}a, respectively. However, upon inspection of the entire provided dataset, we unfortunately felt that the provided reference standard was at best less than pristine, and at worst remarkably inconsistent, especially for IRF (see Figure \ref{fig:chalEx}b). While the reference standard appears markedly more reliable for SRF and PED than for IRF, it is unfortunately leaky in a manner that is likely harmful to supervised learning. The effect of this is that the intensity distributions of fluid pixels are not at all symmetric like those shown in \cite{Chen3D}, due to the encroachment on bright pixels which are not actually fluid. We generated the intensity distributions for the reference standard and verified that they are indeed very different from those shown in \cite{Chen3D}. This puts us in a bind with regard to the challenge, because we have to choose to either live with the ill effects this has on our method, or go through the trouble of correcting the reference standard ourselves but then still get penalized when our output is not similarly leaky. For the results shown in this paper, we opted for the former. 



Our quantitative results are shown in Table \ref{table:quant}. The dice numbers indicate that our approach performed significantly better on the {\it Zeiss} and {\it Heidelberg} devices than on the {\it Topcon} device for SRF and PED. The SRF difference appears less significant in the AVD statistics, but it must be noted that the {\it Topcon} images provided contain significantly less SRF marked in the reference standard ($1.69$ mm$^3$ vs. $6.14$ and $8.95$ for the {\it Zeiss} and {\it Heidelberg} data sets respectively).

\begin{table}
\centering
\caption{Quantitative results in terms of dice index (DI, higher is better) and absolute volume difference (AVD, measured in mm$^3$, given as mean $\pm$ standard deviation, lower is better).}
\begin{tabular}{p{2.25cm} p{2.1cm} p{2.1cm} p{2.1cm} | p{2.1cm}}
\hline
\textbf{Measure} & \textbf{Zeiss} & \textbf{Heidelberg} & \textbf{Topcon} & \textbf{All Devices}\\
\hline
\textbf{DI (IRF)} & 0.537 & 0.478 & 0.547 & 0.522\\
\textbf{DI (SRF)} & 0.671 & 0.781 & 0.483 & 0.682\\
\textbf{DI (PED)} & 0.699 & 0.610 & 0.459 & 0.612\\
\textbf{AVD (IRF)} & $0.248 \pm 0.429$ &$0.296 \pm 0.379$& $0.115 \pm 0.139$ & $0.285 \pm 0.481$\\
\textbf{AVD (SRF)} & $0.089 \pm 0.154$ & $0.103 \pm 0.158$ &$0.073 \pm 0.152$ & $0.115 \pm 0.207$\\
\textbf{AVD (PED)} &$0.174 \pm 0.336$ & $0.086 \pm 0.155$& $0.222 \pm 0.470$ & $0.156 \pm 0.287$\\
\hline
\label{table:quant}
\end{tabular}
\vspace{-0.75cm}
\end{table}

\section{Conclusion}
We presented a deep learning based method for simultaneously and automatically detecting and segmenting intraretinal fluid, subretinal fluid, and pigment epithelial detachment in OCT images of the human retina. We also presented a novel data augmentation method for these images called myopic warping. We obtained a decent performance, despite the use of what we think is imperfect training data. Remarkably, our method did not involve any kind of precise retinal segmentation, and it stands to reason that our method could potentially be improved by adding one to the pre-processing or post-processing steps. We believe that in time, deep learning will prove to be a necessary component to obtaining state-of-the-art results on the automatic fluid segmentation problem.

\end{document}